# Unsupervised Parallel Extraction based Texture for Efficient Image Representation

Mohammed. M. Abdelsamea

Mathematics Department, Assiut University, Egypt

**Abstract.** SOM is a type of unsupervised learning where the goal is to discover some underlying structure of the data. In this paper, a new extraction method based on the main idea of Concurrent Self-Organizing Maps (CSOM), representing a winner-takes-all collection of small SOM networks is proposed. Each SOM of the system is trained individually to provide best results for one class only. The experiments confirm that the proposed features based CSOM is capable to represent image content better than extracted features based on a single big SOM and these proposed features improve the final decision of the CAD. Experiments held on Mammographic Image Analysis Society (MIAS) dataset.

## 1. Introduction

SOM goes about reducing feature space by producing maps of usually one or two dimensions which plot the similarities of the input data by grouping similar data items together. So SOM's accomplish two things, they reduce dimensions, and display similarities. Feature extraction techniques have played an important role in several medical applications. In general, the applications involve the automatic extraction of features from the image which is then used for a variety of classification tasks, such as distinguishing normal tissue from abnormal tissue. Chabat [1] used 13 texture parameters, derived from the histogram, co-occurrence matrix and run-length matrix categories, to differentiate between a variety of obstructive lung diseases in thin-section CT images. Kovalev [2] used texture parameters derived from gradient vectors and from generalized co-occurrence matrices for the characterization of texture of some MRT brain images. Herlidou [3] used texture parameters based on the histogram, co-occurrence matrix, gradient and run-length matrix for the characterization of healthy and pathological human brain tissues (white matter, grey matter, cerebrospinal fluid, tumours and oedema). Mahmoud [4] used the texture analysis approach based on a three-dimensional co-occurrence matrix in order to improve brain tumour characterization. Du-Yih Tsai [5] used four texture features derived from the co-occurrence matrix was used for classification of the heart disease. H.S. Sheshadri [6] used Six textural features for mammogram images derived from the histogram categories was used as a part of developing a computer aided decision system for early detection of breast cancer. Maria-Luiza [7] used texture parameters based on the histogram for tumour classification in mammograms. Feature extraction [8] is a vital component of the Computer Aided Diagnosis (CAD) System [9] that can discriminate between medical tissues to serve as a second reader to aid radiologists. The feature extraction unit is used to prepare data in a form that is easy for a decision support system or a classification unit to use. Compared to the input, the output data from the feature extraction unit is usually of a much lower dimension as well as in a much easier form to classify. Medical images possess a vast amount of texture information relevant to clinical practice [10, 11]. Hence texture is the most promising feature to work on. Texture analysis gives information about the arrangement and spatial properties of fundamental image elements [12]. Coggins [13] has compiled a catalogue of texture definitions. One of the most commonly used texture parameters come from Co-occurrence matrix as a statistical approach [10] which represents texture in an image using properties governing the distribution and relationships of grey-level values in the image methods normally achieve higher discrimination indexes than the structural or transform methods. A statistical discrimination method (fisherfaces algorithm) [14] for feature selection algorithm is used for extracting discriminative information from extracted feature of medical images to be used as inputs to our classification system.

In this paper, CSOM is set up to transform an incoming signal pattern (feature set of images) of arbitrary dimension into a two dimensional lattice by creates a set of Prototype vectors representing the feature set. the proposed features come from prototypes of trained CSOM neural network, in the experiment, these new features are tested in several way and with several extraction method, which proved that it yield highest performance in overall accuracy of classification system.

## 2. Data Collection and Pre-processing Phase

The data collection which has been used in this experiment was taken from the MIAS[15]. The same collection has been used in other studies of automatic mammography classification. Its corpus consists of 322 images, which belong to three big categories: normal, benign and malign. There are 208 normal images, 63 benign and 51 malign, which are considered abnormal. In addition, the abnormal cases are further divided in six categories: microcalcification, circumscribed masses, speculated masses, ill-defined masses, architectural distortion and asymmetry. The pre-processing stage has been done in this paper as same as in[7].

## 3. Feature Extraction

Extracting features by fixed blocs in the image has been considered to be sufficient as an ROI selection method in some medical applications where a large fraction of the image is covered by tissue of interest. In [16], two ROI selection methods, namely, block wise and pixel wise, have proposed previously to give us better texture feature extraction when co-occurrence matrix method is used and hence a more accurate texture feature.

### 3.1. Concurrent Self-Organizing Maps

Concurrent Self-Organizing Maps (CSOM) [17] is a collection of small SOM, which use a global winner-take-all strategy. Each network is used to correctly classify the patterns of one class only and the number of networks equals the number of classes. The CSOM training technique is a supervised one, but for any individual net the SOM specific training algorithm is used. In this method for pattern classification in training patterns sets are built and used the SOM training algorithm independently for each of the n SOMs. The CSOM model for training is shown in Fig.2

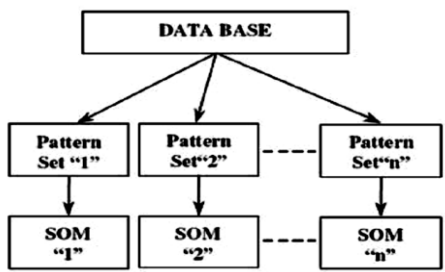
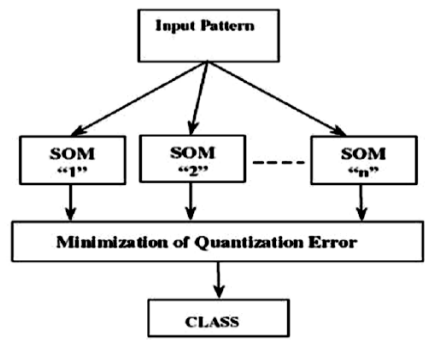

Fig.2: The CSOM model (training phase).      Fig.3: The CSOM model (classification phase).

For the recognition, the test pattern has been applied in parallel to every previously trained SOM. The map providing the least quantization error is decided to be the winner and its index is the class index that the pattern belongs to (see Fig.3).

### 3.1.1. CSOM Extraction Method

In the proposed method, a small SOM for each image features class is set up to transform an incoming signal pattern (feature set of class images) of arbitrary dimension into a two dimensional lattices by creating a set of Prototype vectors representing the features set of each class only. Our proposed features come from prototypes of trained SOM neural network. We apply this extraction method with two different partitioning approaches as a region of interested (ROI) selection methods for extracting different accurate textural features from medical image. Fisherfaces feature selection is used, for selecting discriminated features form extracted textural features. In the experiment, these new features are tested, which proved that it yielded highest performance in overall accuracy of classification system.

The SOM consists of a regular, two dimensional (2-D), grid of map units. Each unit I is represented by a prototype vector WI = {WI1,WI2, . . . ,WIs}, where S is input vector dimension. The units are connected to adjacent ones by a neighbourhood relation. Our proposed features for improved final reduced textural features come from prototypes of trained SOMs network rather than textural features as new features based on old texture features for efficient image representation. In the experiment, we compared our textural feature extracted from the enhancement extraction method with its new features based SOM.

The proposed method is described as, see Figure 3:

**Step 1.** After constructing texture vectors for all images from any of previous three extraction method (TOld) a fisherfaces feature selection is used as image feature data set adaptation for dimensionality reduction to get final reduced textural feature, called textural features set, final set an old textural feature set in the form of transaction TNew = {t1, t2, . . . , tc}, where c is the number of images, every t is a vector of the size S in addition to the class label of the image.

**Step 2.** Divide the whole transaction into number of transaction TCLASS = {t1, t2, . . . , tcn}, where cn is the number of classes.

**Step 3.** For each transaction of divided original transaction TCLASS = {t1, t2, . . . , tcn}, setup small SOM.

**Step 4**: Randomly initialize the weight vector of SOM map.

**Step 5**: Set iteration = 0.

**Step 6:** Select a single transaction of the old textural feature set, TNew = {t1, t2, . . . , tc}.

**Step 7:** Find a winner neuron, the best matching unit (BMU), on the SOM for that input data using similarity Measure

$$\|X - W_I\| = \min_I [\|X - W_I\|]$$

**Step 8**: Set the learning rate and the neighborhood function according to iteration number and update the winner neuron and its neighbors on the SOM map as:

$$W_I(t+1) = W_I(t) + \alpha(t)h_{WI}(t)[X - W_I(t)],$$

where

t : time.

$\alpha(t)$ : Adaptation coefficient.

hWI (t) : Neighborhood kernel centered on the winner unit:

$$h_{WI}(t) = \exp(-\frac{\|r_W - r_I\|^2}{2\sigma^2(t)}),$$

where

$r_W$ and $r_I$ are positions of neurons W and I on the

SOM grid. Both $\alpha(t)$ and $\sigma(t)$ decrease monotonically with time.

**Step 9**: Set Iteration = Iteration + 1.

**Step 10**: If Iteration reaches its maximum value go to next step, otherwise go to Step 5.

**Step 11**: Find the final winner neuron for a single transaction of the old data set and replace feature vector of that transaction with the prototype of its winner neuron as a new feature vector for represent that image. In the experimental the possibility of adding these proposed features to textural feature set is tested rather than replacing.

**Step 12**: Repeat Step 11 for all transaction of the old data set.

**Step 13**: Finally, set a new features based SOM data set in the form of transaction as the input data set TSOM = {t1, t2, . . . , tc}, where c is the number of images, every t is a vector of the size (S) in addition to the class label of the image.

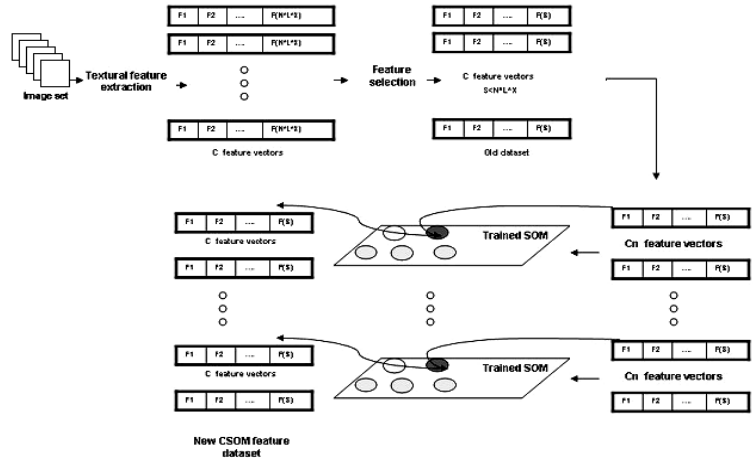

Fig. 3: Generated features based CSOM.

## 4. Experimental Results

In this experiment a sample of 22% from the MIAS dataset is selected randomly for model testing and evaluation. The sample is distributed between class as follows: Normal class (n=30), microcalcification class (n=10), circumscribed masses class (n=7), spiculated masses class (n=8), ill-defined masses class (n=5), architectural distortion class (n=7) and asymmetry class (n=4). The performance of the proposed extraction method is evaluated in terms of overall accuracy with standard classifiers, using Weka experimenter, Such that partitioning our data set using 10-fold cross-validation. Table.1 illustrates comparison between enhancement extraction method [18] based on SOM and CSOM extraction method to accurate image representation based on image feature set extracted using textural extraction method based on textural feature sets based on pixel wise intensity segmentation with SN=6 and L=3. the classification rate comparison with 5 x 5 map size present in fig.4 such that pixel wise segmentation is used.

| Classifiers | CSOM feature 5x5 | Enhancement feature based SOM | | |
|---|---|---|---|---|
| | | 5x5 | 10x10 | 15x15 |
| NNGe | 100% | 88.73% | 100% | 100% |
| J48 | 95.77% | 90.14% | 97.18% | 92.95% |
| Bayes Net | 100% | 91.54% | 100% | 100% |
| NaiveBayes | 100% | 91.54% | 100% | 98.59% |
| Bagging | 100% | 88.73% | 100% | 98.59% |
| RBF Net | 100% | 83.09% | 100% | 100% |

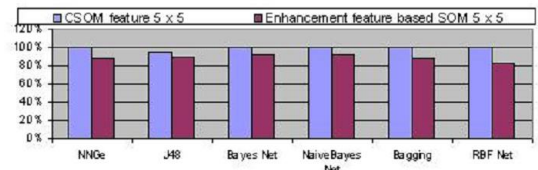

Table.1 the classification rate comparison.          Figure.4: the classification rate comparison.

From the results shown in the table.1, we can notice that, the proposed method gave better accuracy than enhancement extraction method based on pixel wise partitioning as a ROI selection method.

Table.2 illustrates comparison between enhancement extraction method based on SOM and CSOM extraction method to accurate image representation based on image feature set extracted using textural extraction method based on textural feature sets based on our bloc wise segmentation with SN=6, M=8 and L=3.the classification rate comparison with 5 x 5 map size present in fig.5, such that bloc wise segmentation is used.

| Classifiers | CSOM feature 5x5 | Enhancement feature based SOM | | |
|---|---|---|---|---|
| | | 5x5 | 10x10 | 15x15 |
| NNGe | 100% | 98.59% | 100% | 98.59% |
| J48 | 100% | 88.73% | 97.18% | 95.77% |
| Bayes Net | 100% | 88.73% | 100% | 100% |
| NaiveBayes | 100% | 88.73% | 100% | 98.59% |
| Bagging | 100% | 92.95% | 100% | 98.59% |
| RBF Net | 100% | 88.73% | 100% | 100% |

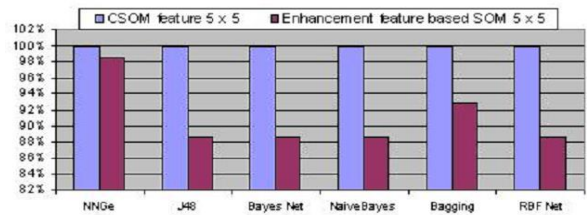

Table.2 the classification rate comparison.          Figure 5: The classification rate comparison.

From the results shown in the table.2, we can notice that, our proposed method gave better accuracy than enhancement extraction method based on Bloc wise partitioning as a ROI selection method.

## 5. Conclusion

In this work, four texture features derived from the co-occurrence matrix was used as part of developing CAD system for early detection of breast cancer. This chapter utilities textural extraction method based on two different ROI selection methods for obtaining efficient image representation. A fisherfaces feature selection algorithm is used for extracting discriminative information from extracted feature of medical images. This paper presents a new extraction method based on prototypes of CSOM network for accurate extraction of textural feature set. The experiments confirm that the proposed features based CSOM is capable to represent image content better than extracted features based on a single big SOM and these proposed features improve the final decision of the CAD.